\documentclass[conference]{IEEEtran}
\IEEEoverridecommandlockouts
\usepackage{cite}

\ifCLASSINFOpdf
\else
\fi
\usepackage{capt-of}
\usepackage{blindtext}
\usepackage{kotex}
\usepackage{graphicx}
\usepackage{array}

\usepackage{tikz}
\usepackage{subfigure}
\usetikzlibrary{positioning,calc}

\usepackage{todonotes}
\usepackage{amsmath}
\usepackage{algpseudocode}
\usepackage{algorithm}
\algnewcommand\algorithmicforeach{\textbf{for each}}
\algdef{S}[FOR]{ForEach}[1]{\algorithmicforeach\ #1\ \algorithmicdo}

\usepackage{dcolumn}
\usepackage{hhline}
\usepackage{multirow}
\newcolumntype{L}[1]{>{\raggedright\let\newline\\\arraybackslash\hspace{0pt}}m{#1}}
\newcolumntype{C}[1]{>{\centering\let\newline\\\arraybackslash\hspace{0pt}}m{#1}}
\newcolumntype{R}[1]{>{\raggedleft\let\newline\\\arraybackslash\hspace{0pt}}m{#1}}

\begin{document}
%
\title{Synonym Discovery with Etymology-based \\ Word Embeddings}

\author{\IEEEauthorblockN{Seunghyun Yoon\textsuperscript{*}}\thanks{*equal contribution}
\IEEEauthorblockA{Department of Electrical and\\Computer Engineering\\
Seoul National University\\
Seoul, Korea\\
Email: mysmilesh@snu.ac.kr}
\and
\IEEEauthorblockN{Pablo Estrada\textsuperscript{*}}
\IEEEauthorblockA{Cloud Dataflow\\Google Inc.\\
Seattle, WA, USA\\
Email: polecito.em@gmail.com}
\and
\IEEEauthorblockN{Kyomin Jung}
\IEEEauthorblockA{Department of Electrical and\\Computer Engineering\\
Seoul National University\\
Seoul, Korea\\
Email: kjung@snu.ac.kr}}

\maketitle

\begin{abstract}

 We propose a novel approach to learn word embeddings based on an extended version of the distributional hypothesis. Our model derives word embedding vectors using the etymological composition of words, rather than the context in which they appear. It has the strength of not requiring a large text corpus, but instead it requires reliable access to etymological roots of words, making it specially fit for languages with logographic writing systems. 

The model consists on three steps: (1) building an etymological graph, which is a bipartite network of words and etymological roots, (2) obtaining the biadjacency matrix of the etymological graph and reducing its dimensionality, (3) using columns/rows of the resulting matrices as embedding vectors.    

We test our model in the Chinese and Sino-Korean vocabularies. Our graphs are formed by a set of 117,000 Chinese words, and a set of 135,000 Sino-Korean words. In both cases we show that our model performs well in the task of synonym discovery.

\end{abstract}


%
\IEEEpeerreviewmaketitle

\section{Introduction}
Word embedding is a very active area of research. It consists of using a text corpus to characterize and embed words into rich high-dimensional vector spaces. By mining a text corpus, it is possible to embed words in a continuous space where semantically similar words are embedded close together. By encoding words into vectors, it is possible to represent semantic properties of these words in a way that is more expressive and useful for tasks of natural language processing. Word embeddings have been effectively used for sentiment analysis, machine translation, and other and other language-related tasks \cite{chen2013expressive, zou2013bilingual}.

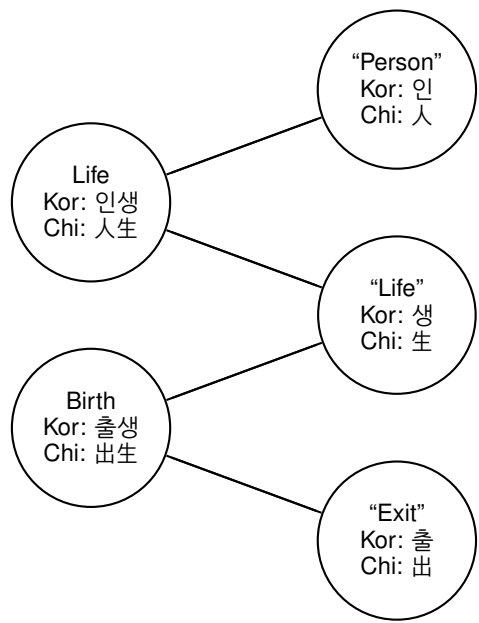
\begin{figure}[htbp]
    \centering
    \begin{tikzpicture}[shorten >=2pt,auto,node distance=3cm,
                    thick,main node/.style={circle,draw,font=\sffamily\small}]
    \node[main node] (1) [draw, text width=1.5cm, align=center] {``Person" \\ Kor: 인 \\ Chi: 人};
    \node[main node] (3) [below of=1,draw, text width=1.5cm, align=center] {``Life" \\ Kor: 생 \\ Chi: 生};
    \node[main node] (2) [left = of {$(1)!0.5!(3)$}, draw, text width=1.5cm, align=center] {Life \\ Kor: 인생 \\ Chi: 人生};
    \node[main node] (4) [below of=2, draw, text width=1.5cm, align=center] {Birth \\ Kor: 출생 \\ Chi: 出生};
    \node[main node] (5) [below of=3,draw, text width=1.5cm, align=center] {``Exit" \\ Kor: 출 \\ Chi: 出};
  \path[every node/.style={font=\sffamily\small}]
    (1) edge node [] {} (2)
    (2) edge node [] {} (1)
        edge node [] {} (3)
    (3) edge node [] {} (2)
        edge node [] {} (4)
    (4) edge node [right] {} (3)
        edge node [right] {} (5)
    (5) edge node [right] {} (4);
\end{tikzpicture}
    \caption{Subset of bipartite graph. On the left are the ``words" that are formed by mixing ``roots", which can be seen on the right. In Sino-Korean vocabulary, most words are formed by using two etymological roots, but there are also words formed by three roots, and a few that are formed by more than three. In Chinese vocabulary, words tend to be longer on average, and thus have a larger amount of etymological roots.
    }
    \label{fig:smgraph}
\end{figure}

The basic idea behind all methods of word embedding is the distributional hypothesis: ``A word is characterized by the company it keeps" \cite{levy2014neural, firth1957synopsis}. Based on this idea, count-based methods such as LSA \cite{dumais2004latent}, and predictive methods that use neural networks to learn the embedding vectors were developed, and used in research with success \cite{mikolov2013efficient,baroni2014don}.

In this work, we propose a new approach to learn word embeddings that is based on the etymological roots of words. Our approach relies on the fact that a shared etymological root between two words expresses a deliberate semantic similarity between these two words; by leveraging information on these semantic similarities, we derive the embedding vectors of words. This is akin to extending the distributional hypothesis to consider etymological context as well as textual context: words that appear in similar \textit{etymological} contexts must also express similar concepts.

Based on this hypothesis, our approach consists of building a graph that captures these etymological relationships, and reducing the dimensionality of its adjacency matrix to learn word embeddings. Our approach can be applied to vocabularies of any language. Since our work relies on etymology, it requires a dependable way to obtain the etymological roots of words. This is why our approach is particularly well suited for the Chinese language, and other languages that borrow some vocabularies from Chinese. Chinese characters are representative ideograms, so that we can consider each characters as the etymological information which are available and easily accessible. We note that our approach can be also applied to other languages with known etymological roots of words, for example, English or Spanish with Latin root of words.

To verify the word embeddings learned by our model we use the task of synonym discovery, whereby we analyze if it is possible to identify a pair of words as synonyms only through their embedding vectors. Synonym discovery is a common task in research; and it has been used before to test word embedding  schemes \cite{chen2013expressive}. We compare the performance of our Chinese word embedding vectors in the task of synonym discovery against another set of embedding vectors that was constructed with a co-occurrence model\cite{zou2013bilingual}. We also investigate the performance of synonym discovery with the Sino-Korean word embeddings by our method. Our test results shows that our approach out-performs the previous model.

Our approach can be applied to vocabularies of any language. Since our work relies on etymology, it requires a dependable way to obtain the etymological roots of words. In languages with primarily phonetic writing systems, inferring the etymological roots of words is a significant challenge that requires intellectual work to trace words back to their ancestors. This is perhaps the reason that not much research has been made in the data mining community that is based on etymology. That stands in contrast to languages with logographic writing systems, where a word carries morphological information in its writing. This makes the task of etymology extraction much simpler. This is why our approach is particularly well suited for the Chinese language, and the subset of the Korean vocabulary that is comprised by Sino-Korean words (i.e. Korean words that have been borrowed from Chinese).

Written Chinese is comprised by a large set of \textit{Hanzi}, or characters. Generally, one character represents one syllable of spoken Chinese; and it may represent a monosylabic word, or be part of a polysyllabic word. The characters themselves can be composed to form new, more complex, characters. Chinese writing has also been adopted in other languages such as Korean, Japanese and formerly also Vietnamese. In this work, we use each character as an \textit{etymological root} that forms part of a word (which is either mono- or polysyllabic); and we study Chinese vocabulary in Korean and in the Chinese language.

\section{Related work}
There exists limited research on etymological networks in the English language. Particularly \cite{hunter2015network}, and \cite{hunter2014novel} use an etymological network-based approach to study movie scripts and reviews in English.

When it comes to work that studies the Chinese writing system, a popular topic is to study how radicals combine to form more complex characters \cite{li2007chinese}. Some studies have created networks based on word co-occurrence \cite{zhou2008empirical}. We found only one study that creates a network based on how characters mix to form words \cite{yamamoto2009network}.

The task of synonym discovery in Chinese vocabulary has been tackled in previous work \cite{yong2008research,lu2009using}. These studies use a large corpus from the Chinese Wikipedia, and identify synonyms by building a graph where words are linked if their Wikipedia articles link to each other. These studies do not report their performance in general, instead reporting some identified synonym pairs.

In another study, \cite{pablo2016knowledge} defined an etymological graph-based framework from Sino-Korean data, and used it in a supervised classification scheme to find pairs of Chinese characters (e.g. etymological roots) that were synonyms. It showed that the etymological graph approach can be effectively used to extract knowledge from a complex etymological network.

Word embedding was defined originally in \cite{bengio2006neural}, where the authors use a neural network-based approach to generate a language model of which word embeddings are a byproduct. Since then, numerous studies have been written where both neural networks and count-based models have been used to produce word embeddings \cite{mikolov2013efficient, baroni2014don}. Aligned embeddings have also been used for machine translation, particularly \cite{zou2013bilingual} attempts translation between English and the Chinese language. 

To the best of our knowledge, there are no papers that explore any data mining task based on etymology in either languages with phonetic alphabets or with logographic alphabets.


\section{Method}

\begin{algorithm}[t]
\small
\caption{Building etymological graph}
\label{alg:build_graph}
\begin{algorithmic}[1]
\Require Empty graph $ \mathcal G=(\mathcal V,\mathcal E) $
\Require List of words $ \mathcal W $ annotated with etymological roots.
\ForEach {$w \in \mathcal W $}
\State $\mathcal V \gets \mathcal V \bigcup \{w\}$
\ForEach {$root \in w $}
\If {$root \notin \mathcal V$} 
\State $\mathcal V \gets \mathcal V \bigcup \{root\}$
\EndIf
\State $ \mathcal E \gets \mathcal E \bigcup \{\{root,w\}\}$
\EndFor
\EndFor
\end{algorithmic}
\end{algorithm}

\begin{figure*}[ht]
\centering
\subfigure[Chinese vocabulary]{\label{fig:chRvsS}\includegraphics[width=1\columnwidth]{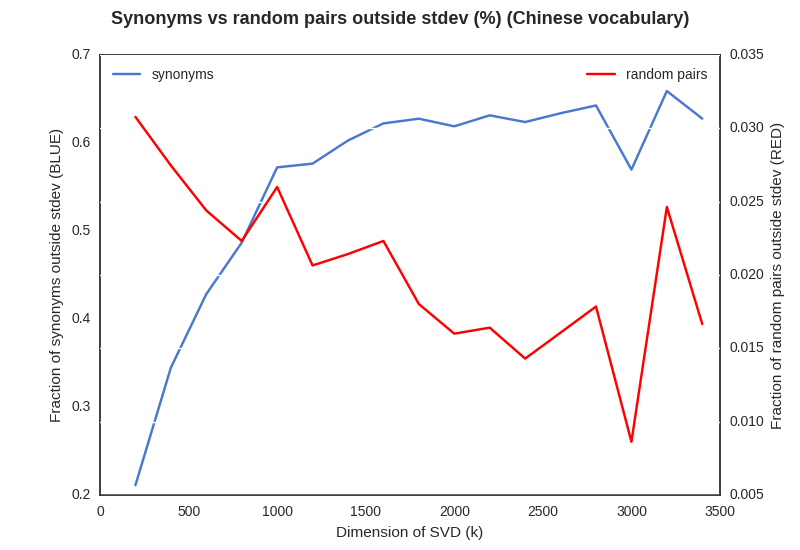}}
\subfigure[Sino-Korean vocabulary]{\label{fig:krRvsS}\includegraphics[width=1\columnwidth]{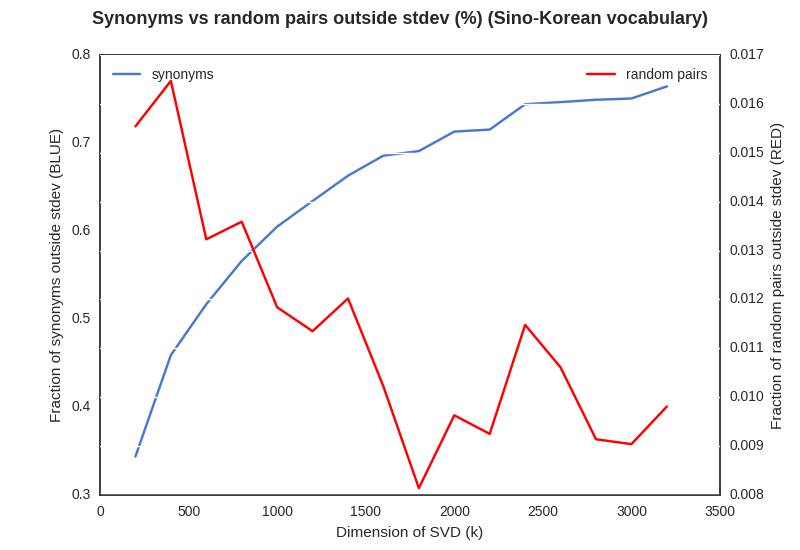}}
\caption{Proportion of random pairs of words where the dot product of their embedding vectors is far from zero (RED) and proportions of pairs of synonyms where the dot product of their embedding vectors is far from zero (BLUE). Only 1\% of the random pairs place far from zero, while about 65\%(a) and 73\%(b) of synonyms are far from zero.}
\label{fig:proportions}
\end{figure*}


\subsection{Building the etymological graph}
An etymological graph is a bipartite network with two sets of nodes: one that represents the roots of the words in a language, while the other set represents the words themselves. In an etymological graph, two nodes are connected if one node represents an etymological root of the word represented by the other, as shown in Figure \ref{fig:smgraph}.

To build an etymological graph, one may start from a list of words annotated with their etymological roots. By iterating over the list, and iterating over the roots of each word; it is possible to add nodes and edges to the graph in order. This procedure is expressed in algorithm \ref{alg:build_graph}.

\begin{table}[t]
\caption{Statistics from the Sino-Korean vocabulary graph, and the Chinese vocabulary graph.}
\label{tab:graphstats}
\centering
\begin{tabular}{|C{0.40\columnwidth}|C{0.18\columnwidth}|C{0.18\columnwidth}|}
\hline
    \textbf{Language} & \textbf{Chinese} & \textbf{Sino-Korean} \\ \hline
    Num. of Words & 117,568 & 136,045 \\
    Num. of Characters & 5,115 &  5,972\\
    Avg. word length & 3.36 & 2.56\\
    Avg. degree of a root-node & 76.45 & 58.2 \\ 
    \hhline{|===|}
    \multicolumn{3}{|c|}{\textbf{Words by length}} \\ \hline
    1 character & 2,082 & - \\
    2 characters & 25,001 & 77,891\\
    3 characters & 35,108 & 40,024 \\
    4 characters & 39,249 & 18,130 \\
    5 characters & 16,128 & - \\ \hline
\end{tabular}
\end{table}

As part of our research, we built two graphs using data collected by crawling an online dictionary for the set of Sino-Korean vocabulary\footnote{Available from ``http://hanja.naver.com/"}; and online Chinese dataset for Chinese vocabulary\footnote{Available from ``http://adsotrans.com/downloads/''}. Some statistics about these graphs are shown on table \ref{tab:graphstats}. It is interesting to note that the distributions over word length in Chinese is different than in Korean. This is, perhaps, due to the differences in the ways Chinese loan-words are used in the Korean language, and the ways Chinese uses its own words. These differences should not affect the outcome of our model, because they do not affect the construction of the graph.

\subsection{Learning word embeddings}
To obtain the word embeddings from the graphs, truncated Singular Value Decomposition (SVD) was applied to their biadjacency matrices \cite{asratian1998bipartite}. We use SVD inspired by the techniques of LSA \cite{dumais2004latent}, where it is possible to map words and documents to ``hidden" semantic characteristics.

The bi adjacency matrix $A$ of a bipartite graph is a matrix of size $n \times m$ where each column represents a node from one bipartite set, and each row represents a node from the other bipartite set. In the case of etymological graphs, each row represents a root node, while each column represents a word node; therefore the matrix $A$ has dimension $\#roots \times \#words$. 

By applying SVD, we attempt to approximate the biadjacency matrix $ A $ as the product of three matrices $ U \Sigma V^*$, which is the closest $k$-dimension approximation of $A$. In this operation, $\Sigma$ is a diagonal matrix with the $k$ largest singular values in the diagonal, and the matrices $U$ and $V^*$ are matrices of size $\#roots \times k$ and $k \times \#words$ respectively; where $k$ is the dimension into which we chose to reduce matrix $A$.  We use the dimension-reduced column vectors in $V^*$ as embeddings for each word in our vocabulary.

Another matrix decomposition technique worth considering for future work is CUR factorization \cite{boutsidis2017optimal}. We're specially interested in its sparsity-maintaining characteristic; since large matrices such as ours can be managed more easily if they are sparse - and SVD eliminates the sparsity of our source matrices.

\subsection{Verifying the word embeddings: Synonym discovery}
To verify the validity of the embeddings, we selected the task of synonym discovery. To assess whether two words are synonyms, we measure their cosine similarity (e.g. internal product) as proposed in \cite{blondel2004measure}. We expect synonyms to show similarity score above a threshold, which we decide by studying the distribution of the similarity between random pairs of words. In other words, we obtain the dot product of vectors from random pairs of words, and compare them to the dot product of vectors from pairs of synonyms. As random pairs of words are expected to have little semantic relationship, the dot product of their embedded vectors is expected to be close to 0; while the dot product of vectors representing pairs of synonyms is expected to be far from 0 due to the semantic similarity between pair of synonyms, which should be expressed by their embedding in a vector space.


\begin{figure*}[th!]
\centering
\subfigure[k=380]{\label{fig:krDotProd380}\includegraphics[width=1.0\columnwidth]{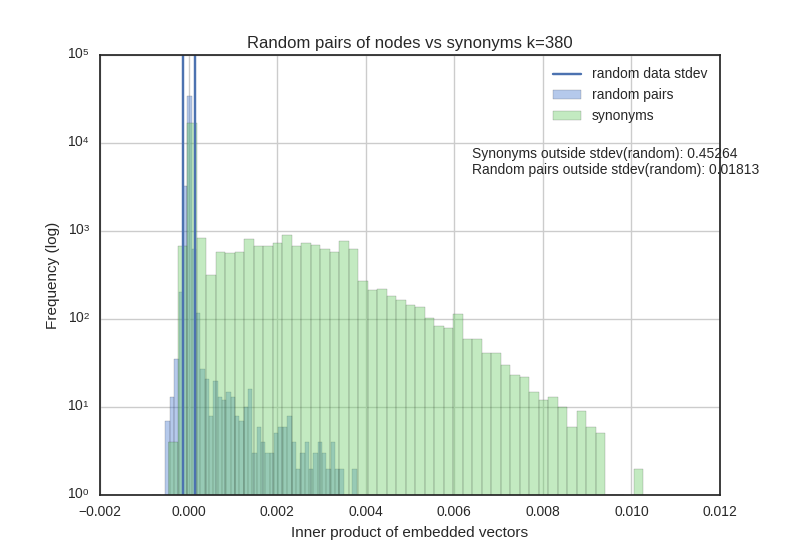}}
\subfigure[k=740]{\label{fig:krDotProd740}\includegraphics[width=1.0\columnwidth]{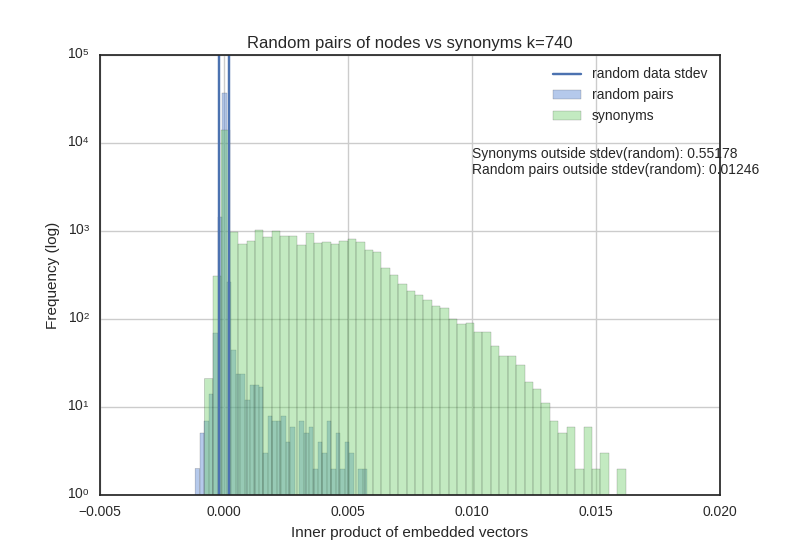}}
\subfigure[k=1100]{\label{fig:krDotProd1100}\includegraphics[width=1.0\columnwidth]{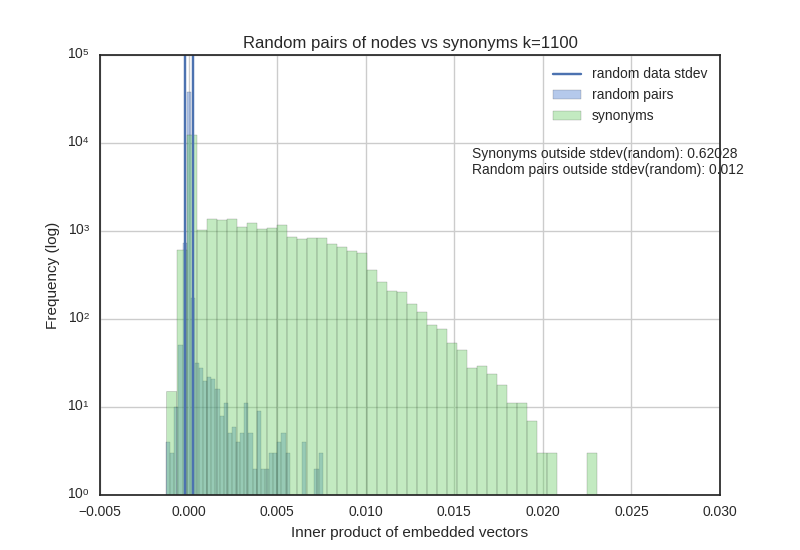}}
\subfigure[k=1640]{\label{fig:krDotProd1640}\includegraphics[width=1.0\columnwidth]{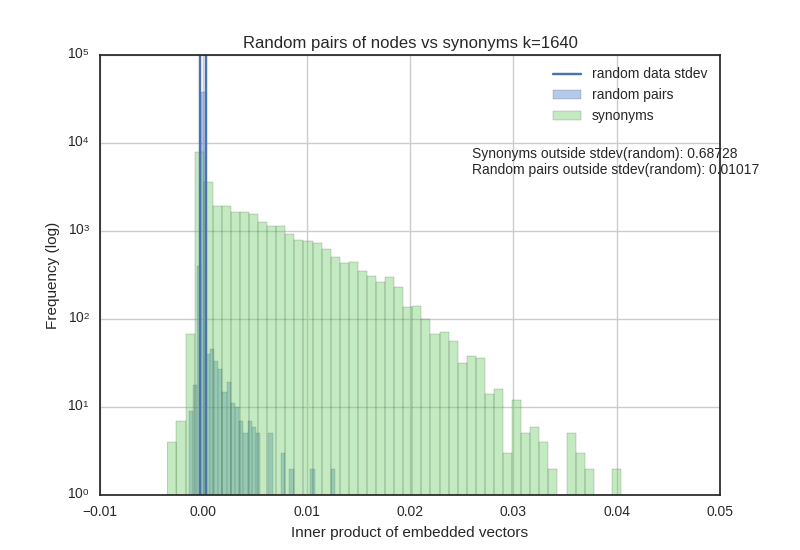}}
\subfigure[k=1820]{\label{fig:krDotProd1820}\includegraphics[width=1.0\columnwidth]{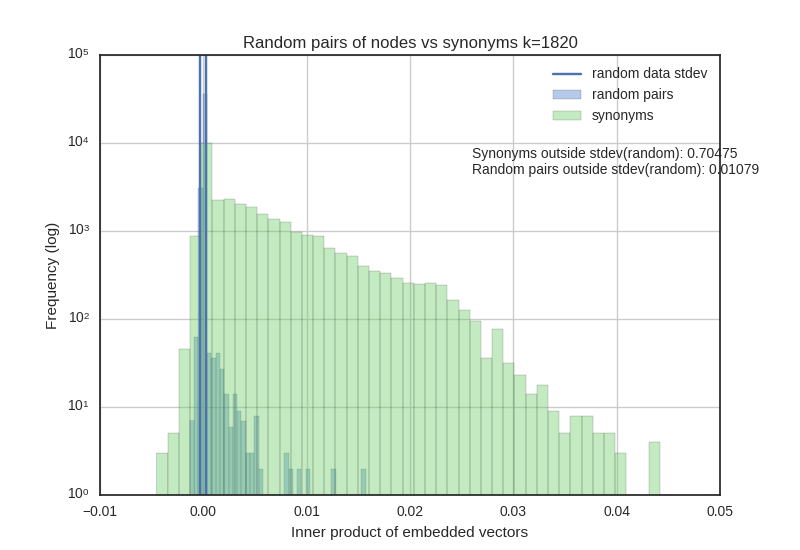}}
\subfigure[Result from embeddings released from \cite{zou2013bilingual}]
{\label{fig:other_hist}\includegraphics[width=1.0\columnwidth]{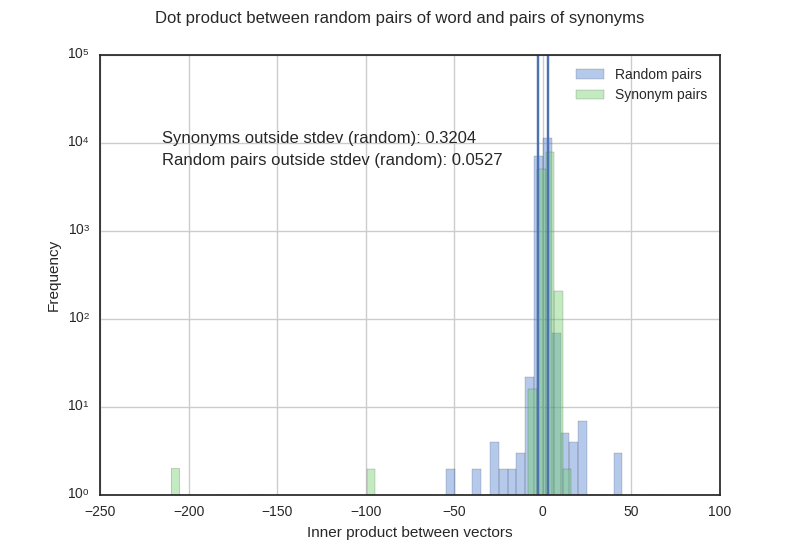}}
\caption{Log-scale histograms of dot product between embedding vectors between (green) pairs of synonyms, and (blue) random pairs of words. The vertical lines (blue) represent the standard deviation of the distribution of dot products between random pairs of words.}
\label{fig:histograms6}
\end{figure*}

\section{Experiments}
For comparison, we used the dataset of Chinese word embeddings that was released as part of  \cite{zou2013bilingual}, which contains embeddings of Chinese words in 50 dimensions. 
We used this data set on the same task: Synonym discovery by measuring their similarity score as the internal product between vectors.


\begin{table}[ht!]
	\caption{Model Performance in Classifying Synonyms}
    \label{tab:all_models}
	\centering
	\begin{tabular}{|C{0.14\columnwidth}|C{0.12\columnwidth}|C{0.12\columnwidth}|C{0.18\columnwidth}|C{0.18\columnwidth}|}
    \hline
        \textbf{Model} & \textbf{Language} & \textbf{Dimensions} & \textbf{Correctly classified synonyms} & \textbf{Misclassified random pairs} \\ 
        \hline
        Zou, \textit{et al.} & Chinese & 50 & 32\% & 5\% \\
        \hhline{|=====|}
        Our model & Korean & 2000 & 70\% & 1\% \\
        Our model & Chinese & 2000 & 64\% & 1.5\% \\
         
        \hline
    \end{tabular}
\end{table}


To obtain the ``ground truth" of synonym pairs, we collected pairs of synonyms from online dictionaries  for both Chinese and Sino-Korean vocabulary. For Sino-Korean, we generated query with the words from Sino-Korean vocabulary and crawled synonyms by searching data on the web \cite{Naver_dic}. In the same way, we crawled synonyms of Chinese vocabulary by searching data on the Chinese Synonyms Thesaurus service \cite{Chinese_synonyms_thesaurus}. With this way we collected a total of 38,593 pairs of synonyms, while in Chinese we collected 45,731 pairs.


\subsection{Performance of synonym discovery task}
Our experiments show that we were able to reliably identify pairs of synonyms by comparing the embeddings of pairs of words. Performance was specially good in the Korean language graph, as can be seen in Figure \ref{fig:proportions}, where we plot distributions of dot product between random pairs and pairs of synonyms. As shown in the figure, up to 70\% of all synonyms have a similarity measure that places them outside the range covered by 99\% of random pairs of synonyms. Figure \ref{fig:histograms6} helps drive this point by showing the variation of the proportion of synonyms that are placed outside the standard deviation of the distribution of dot products of embeddings of random pairs of words when we vary the dimension of our embeddings. Note how only about 1\% of random pairs of words appear outside of this range, and the vast majority of them consistently concentrated around zero.

Our embeddings also proved to perform better than our benchmark dataset. Figure \ref{fig:other_hist} shows the distribution of the similarity measure between pairs of synonyms and random pairs of words in the benchmark dataset. In this sample, almost 32\% of synonyms show a similarity score that places them away from zero, while 5\% of random pairs of words are placed outside of that range. Table \ref{tab:all_models} compares performance, and dimensionality in both strategies to learn embeddings.

\subsection{Computation speed of our model}
An interesting feature of word embedding models based on matrix factorizations is that training time can be significantly shorter when compared with the time it may take to train a multi-layered neural network. For dimensions under 500, SVD can run very quickly, but as the dimension rises, the factorization step becomes significantly slower. Our model reaches its best performance at around 2000 dimensions, for which the matrix factorization takes over 5 minutes of computation.

Code for our model was developed in Python 3. Paticularly, we used the NetworkX python package to manage and analyze our graphs, and the SciPy and the NumPy  libraries to work with matrices and vectors ~\cite{hagberg2008exploring, jones2014scipy, walt2011numpy}. Our code ran on a Intel Core i7-4790 clocked at 3.60GHz and 16 GB of RAM. table \ref{tab:runtime} shows the running time of the factorization of both our graphs and different values for the dimension of the matrix decomposition.


\begin{table}[t]
    \caption{Running time of matrix factorization}
    \label{tab:runtime}
    \centering
	\begin{tabular}{|c|c|c|}
    \hline
    \multirow{2}{*}{\textbf{SVD k}} & \multicolumn{2}{c|}{\textbf{Time (seconds)}}    \\ \cline{2-3}
                           & \textbf{Chinese vocabulary} & \textbf{Sino-Korean vocabulary} \\
    \hline                       
    200                    & 4.6                & 6.3                    \\
    600                    & 29.2               & 36.7                   \\
    1000                   & 56.9               & 71.3                   \\
    1400                   & 113.36             & 144.9                  \\
    1800                   & 215.7              & 266.2                  \\
    2200                   & 312.4              & 407.2                  \\
    2400                   & 337.2              & 484.6                  \\
    2600                   & 346.2              & 566.4                  \\
    3000                   & 369.1              & 737.4					 \\
    \hline
    \end{tabular}
\end{table}

These running times stand in contrast with the rather large times it takes to train a neural network model. Nonetheless, given that our embeddings require a higher number of dimensions to be effective, SVD on the dimension that we require has a relatively slow performance of up to 5 minutes.

The code for this paper, as well as the datasets and instructions on how to replicate this work are openly available \cite{HanjaGraph}.

\section{Conclusion}
In this work, we have presented a model to learn word embeddings based on etymology. We have shown that the model can capture semantic information derived from a complex etymological network. Its performance is remarkably good in the task of synonym discovery. We believe it can also perform well in other tasks such as antonym discovery.

A noticeable difference between our word embeddings and existing ones is that ours require a much higher number of dimensions to perform well in synonym discovery. Publicly available datasets with word embeddings provide vectors with 25, 50 and 100 dimensions ~\cite{chen2013expressive}; but our embeddings reach their highest effectiveness at around 2,000 dimensions. This is likely a consequence of our data being very sparse: while words in word co-occurrence models can have an almost limitless set of contexts in which they appear, words in etymological graphs have a small number of etymological roots. All the words in our graphs are formed by 5 characters or less. 



The approach covered in this paper also has some particular quirks that stem from the use of \textit{historical} (i.e. etymological) data. This is because the meaning of words is not static, but rather evolves with time and use. Word embeddings that are learned from co-occurrence models are able to capture the ways in which words are used in target corpus.This is not captured by a top-down model that is based on etymology. Our approach would capture the semantics of words from the word roots, rather than how they are used in the text.

Our model also does not rely on very large text corpora, though instead it requires reliable access to etymological roots of words. Etymological dictionaries already capture some of this data, but languages continue to evolve and words to be coined at an ever faster pace, so techniques of machine learning will have to be used to obtain reliable access to etymological roots in other languages.

We believe that our model can help expand our understanding of word embedding; and also help reevaluate the value of etymology in data mining and machine learning. We are excited to see etymological graphs used in other ways to extract knowledge. We also are especially interested in seeing this model applied to different languages.

\section*{Acknowledgment}
K. Jung is with the Department of Electrical and Computer Engineering, ASRI, Seoul National University, Seoul, Korea. This work was supported by Basic Science Research Program through the National Research Foundation of Korea(NRF) funded by the Ministry of Education(NRF-2016M3C4A7952587), the Ministry of Trade, Industry \& Energy(MOTIE, Korea) under Industrial Technology Innovation Program(No.10073144), and the Brain Korea 21 Plus Project.



%


\bibliographystyle{IEEEtran}
\bibliography{./IEEEabrv,./SSCI2017}

\end{document}